\documentclass[letterpaper, 10 pt, conference]{ieeeconf}  
\usepackage{algorithm}
\usepackage{algorithmic}
\usepackage{graphicx}
\usepackage{xcolor}
\usepackage{amssymb}
\usepackage{amsmath}
\usepackage{multirow}
\usepackage[normalem]{ulem}
\graphicspath{{./Images/}}
\makeatletter

\newcommand\fs@norules{\def\@fs@cfont{\bfseries}\let\@fs@capt\floatc@ruled
  \def\@fs@pre{}%
  \def\@fs@post{}%
  \def\@fs@mid{\kern3pt}%
  \let\@fs@iftopcapt\iftrue}
\makeatother
\floatstyle{norules}
\restylefloat{algorithm}

\IEEEoverridecommandlockouts                              
    
\title{\LARGE \bf
Learning Coordination Policies over Heterogeneous Graphs for Human-Robot Teams via Recurrent Neural Schedule Propagation
}
\author{Batuhan Altundas$^{1}$, Zheyuan Wang$^{1}$, \textcolor{black}{Joshua Bishop}$^{1}$ and Matthew Gombolay$^{1}$
\thanks{*This work was supported in part by the Office of Naval Research under grant \textcolor{black}{N00014-19-1-2076} and \textcolor{black}{Naval Research Laboratory} under grant \textcolor{black}{N00173-21-1-G009}.}
\thanks{$^{1}$Batuhan Altundas, Zheyuan Wang, \textcolor{black}{Joshua Bishop} and Matthew Gombolay are with the Institute for Robotics and Intelligent Machines, Georgia Institute of Technology, Atlanta, GA 30332, USA
        {\tt\small \{baltundas3, pjohnwang, \textcolor{black}{jbishop45}, mgombolay3\}@gatech.edu}}%
}
\begin{document}

\maketitle
\thispagestyle{empty}
\pagestyle{empty}

\begin{abstract}
As human-robot collaboration increases in the workforce, it becomes essential for human-robot teams to coordinate efficiently and intuitively. Traditional approaches for human-robot scheduling either utilize exact methods that are intractable for large-scale problems and struggle to account for stochastic, time varying human task performance, or  application-specific heuristics that require expert domain knowledge to develop. We propose a deep learning-based framework, called HybridNet, combining a heterogeneous graph-based encoder with a recurrent schedule propagator for scheduling stochastic human-robot teams under upper- and lower-bound temporal constraints. The HybridNet's encoder leverages Heterogeneous Graph Attention Networks to model the initial environment and team dynamics while accounting for the constraints. By formulating task scheduling as a sequential decision-making process, the HybridNet's recurrent neural schedule propagator leverages Long Short-Term Memory (LSTM) models to propagate forward consequences of actions to carry out fast schedule generation, removing the need to interact with the environment between every task-agent pair selection. The resulting scheduling policy network provides a computationally lightweight yet highly expressive model that is end-to-end trainable via Reinforcement Learning algorithms. We develop a virtual task scheduling environment for mixed human-robot teams in a multi-round setting, capable of modeling the stochastic learning behaviors of human workers. Experimental results showed that HybridNet outperformed other human-robot scheduling solutions across problem sizes for both deterministic and stochastic human performance, with faster runtime compared to pure-GNN-based schedulers.
\end{abstract}

\section{INTRODUCTION}

With collaborative robots (cobots) becoming more available in the industrial and manufacturing environments, robots and humans \textcolor{black}{increasingly share} the same work space \textcolor{black}{to collaborate on tasks} \cite{Yan2013}. By removing the cage around traditional robot platforms and integrating \textcolor{black}{robots into} dynamic, final assembly \textcolor{black}{operations, manufacturers} can see improvements in reducing a factory's footprint and environmental \textcolor{black}{costs as} well as increased productivity \cite{heyer2010human}. In this paper, we focus on the problem of multi-agent task allocation and scheduling \cite{nunes2017taxonomy} with mixed human-robot teams over multiple iterations of the same task allocation problem. 
\textcolor{black}{Our work accounts for and leverages stochastic, time-varying human task performance to quickly solve task allocation problems among team members to achieve a high-quality schedule with respect to the application-specific objective function while satisfying the temporal constraints (i.e., upper and lower bound deadline, wait, and task duration constraints).}



Compared to task scheduling within multi-robot systems, the inclusion of human workers makes scheduling even more challenging because, while robots can be programmed to carry out certain tasks at a fixed rate, human workers typically have latent, dynamic, and task-specific proficiencies. Effective collaboration in human-robot teams requires utilizing the distinct abilities of each team member to achieve safe, effective, and fluent execution. For these problems, we \textcolor{black}{must} consider the ability of humans to learn and improve \textcolor{black}{in task performance} over time.
\textcolor{black}{To exploit this property, a scheduling algorithm must reason about a human's latent performance characteristics in order to decide whether to assign the best worker to a task now versus giving more task experience to a person who is slower but has a greater potential for fluency at that particular task.}
However, it is non-trivial to infer human strengths and weaknesses while ensuring that the team satisfies requisite scheduling constraints, due to the uncertainty introduced by variability in task execution behavior across different individuals, as well as uncertainty on future task performance affected by human's learning effects with practice \cite{humanrobotteam}. Moreover, a lack of consideration for human preferences and perceived equality may, in the long run, put efficient behavior and fluent coordination at a contradiction \cite{gombolay2015decision}.


Recent advances in scheduling methods for human-robot teams have shown a significant improvement in the ability to dynamically coordinate large-scale teams in final assembly manufacturing \cite{nunes2015multi, gombolay2018fast}. Prior approaches typically rely on an assumption of deterministic or static worker-task proficiencies to formulate the scheduling problem as a mixed-integer linear program (MILP), which is generally NP-hard \cite{solomon1986worst}\textcolor{black}{. Exact} methods are hard to scale and often fail to consider the time-varying stochastic task proficiencies of human workers over multi-round schedule execution that could result in significant productivity gains. 
\textcolor{black}{The heuristic approaches} may be able to determine task \textcolor{black}{assignments; however, such approaches required domain specific knowledge that takes years to gain.}
We desire a scalable algorithmic approach that can automatically learn to factor in the human behavior for fast and fluent human-robot teaming.


\begin{figure*}[thpb]
      \centering
    \framebox{\includegraphics[width=0.8\textwidth]{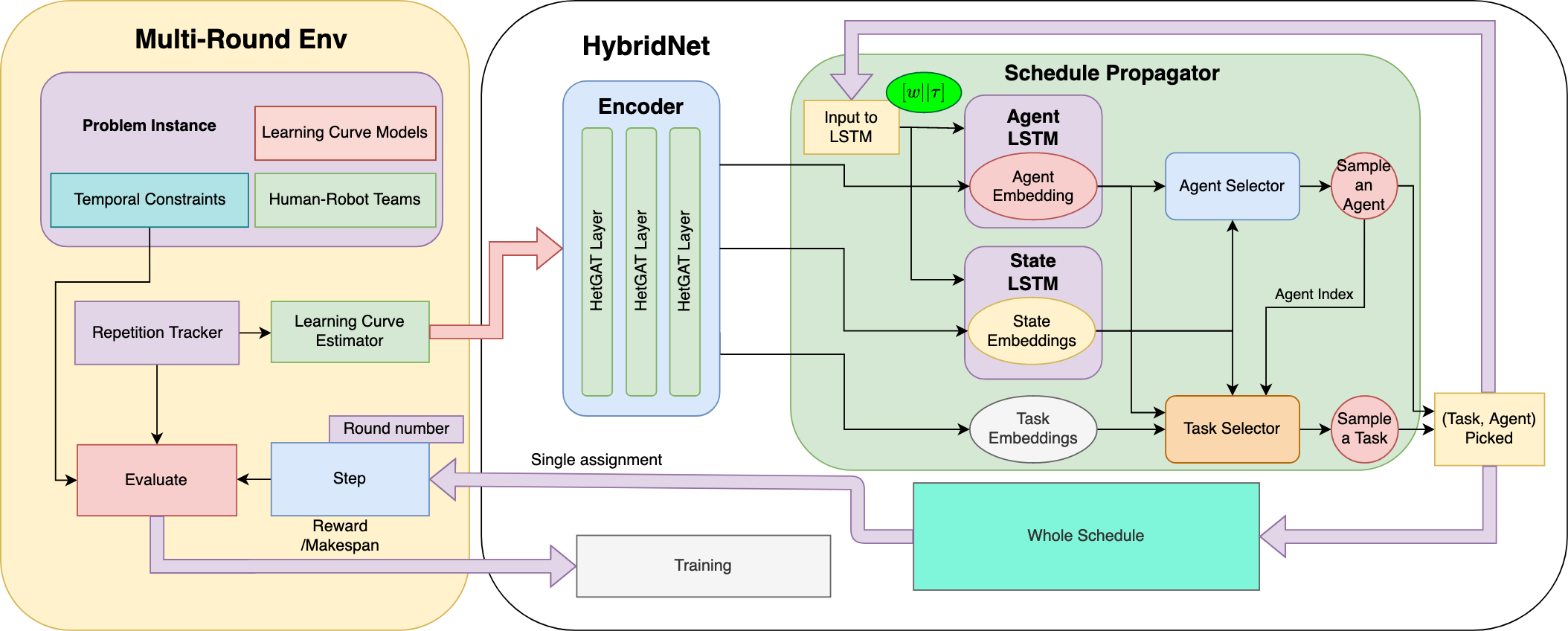}}
    \caption{Overview of Multi-Round Environment with HybridNet Scheduler. Left: The Multi-Round Scheduling Environment is developed to simulate a human-robot scheduling problem over multiple iterative rounds of execution, accounting for changes in human task performance. Right: HybridNet consists of a heterogeneous graph-based encoder to extract high-level embeddings of the problem and a recurrent schedule propagator for fast  schedule  generation.}
    \label{fig:pipeline}
\vspace{-5mm}
\end{figure*}

Advancements in artificial intelligence have fostered the idea of leveraging deep neural networks (DNNs) to solve a plethora of problems in operations research \cite{bengio2021machine}. DNNs can be trained to automatically explore the problem structure and discover useful representations in high-dimensional data towards constructing high-quality solutions, without hand-crafted feature engineering \cite{lecun2015deep}. Particularly, promising progress has been made in learning scalable solvers with graph neural networks via imitation learning (IL) or reinforcement learning (RL), outperforming state-of-the-art, approximate methods \cite{khalil2017learning, kool2018attention, Ma_Ferber_Huo_Chen_Katz_2020}\textcolor{black}{.}


To overcome the limitations of prior work, we propose a deep learning-based framework, called HybridNet, for scheduling stochastic human-robot teams under temporal constraints. \textcolor{black}{Figure} \ref{fig:pipeline} shows the overall framework of our proposed method operating in a multi-round environment. HybridNet utilizes a heterogeneous graph-based encoder and a recurrent schedule \textcolor{black}{propagator}. The encoder extracts high level embeddings of the scheduling problem using a heterogeneous graph representation of the problem extended from the simple temporal network (STN)~\cite{dechter1991temporal}. By formulating task scheduling as a sequential decision-making process, the recurrent \textcolor{black}{propagator} uses Long Short Term Memory (LSTM) cells to carry out fast schedule generation. The resulted policy network provides a computationally lightweight yet highly expressive model that is end-to-end trainable via reinforcement learning \textcolor{black}{algorithms}.

The primary contributions of our work are:
\begin{itemize}
\item We propose a deep learning-based framework, HybridNet, for human-robot coordination under temporal constraints. HybridNet consist of a Heterogeneous Graph-based encoder and a Recurrent Schedule Propagator. The encoder extracts relevant information about the initial environment, while the Propagator generates the consequential models of each task-agent assignments based on the initial model. Inspired by the sensory encoding and recurrent processing of the brain, this approach allows for fast schedule generation, removing the need to interact with the environment between every task-agent pair selection.

\item We develop a virtual task scheduling environment for mixed human-robot teams in a multi-round setting, capable of modeling the stochastic learning behavior of human workers. We make our environment OpenAI gym-compatible and expect it to serve as a testbed to facilitate the development of human-robot scheduling algorithms. \textcolor{black}{The implementation is publicly available.\footnote{https://github.com/altundasbatu/HybridNet\_IROS2022}}

\item \textcolor{black}{We present a novel policy model that \textcolor{black}{jointly} learns how to pick agents and tasks without interacting with the environment \textcolor{black}{between intermediate scheduling decisions and only needs a single reward at the end of schedule.} By factoring in} the action space into an agent selector and a task selector, \textcolor{black}{we} enable conditional policy learning with HybridNet. We account for the state and agent models when selecting the agents, and combine the information regarding the tasks, selected agent and the state for task assignment. \textcolor{black}{As a result,} HybridNet \textcolor{black}{is} end-to-end trainable via Policy Gradients algorithms.

\item We conducted extensive experiments to validate HybridNet across a set of problem sizes. Results showed HybridNet consistently outperformed prior human-robot scheduling solutions \textcolor{black}{under both deterministic and stochastic settings.} 





\end{itemize}

\section{RELATED WORK}

\subsection{Multi-Agent Scheduling Problem}

Task assignment and scheduling of multi-agent systems is an optimization problem that has been studied for real world applications, both for Multi-Robot Task Allocation(MRTA) problem using traditional techniques \cite{Nunes} and deep \textcolor{black}{learning} based techniques \cite{learning_scheduling_with_gnn} as well as for human-robot collaboration \cite{gombolay2018fast}. Task Allocation can be formalised by Mixed Integer Linear Programming (MILP) to capture it's constraints. The exponential complexity of solving the MILP can be accelerated through constraint programming methods \cite{gombolay2018fast, Benders1962, ren2009} or heuristic schedulers to leverage better scalability \cite{Castro2011, chen2009}. Zhang et al. encoded task schedules as chromosomes for a genetic algorithm that optimized schedules for heterogeneous human-robot collaboration by repeatedly crossing over and mutating the solutions to find the optimal schedule. \cite{Zhang2020RealTimeAA}

Gombolay et al. present an algorithm to capture domain knowledge through scheduling policy requiring domain-expert demonstrations \cite{apprenticeship_scheduling}. Wang et al. propose ScheduleNet, a Heterogenous Graph Neural Networks-based model for task allocation under temporospatial constraints, trained through Imitation Learning using optimal schedule \cite{schedulenet}. ScheduleNet relies on interactive scheduling scheme, with constant update of an environment \textcolor{black}{before reaching a complete schedule}. These approaches require optimal schedules generated by other expert systems\textcolor{black}{ to train and} have high computational complexity that make their implementation costly.

\subsection{Modeling Human-Robot Teams}
As advancements in robot capability progress, they become safer and effective to use in conjunction with humans to complete specialized works. \textcolor{black}{Liu et al. presents a model of \textcolor{black}{human task completions}, showing an increase in the \textcolor{black}{task efficiency as a result of learning. This paper shows} that prediction of human performance enhances the ability of the scheduling systems to explicitly reason about the \textcolor{black}{agents'} capabilities \cite{humanrobotteam}}. Prior work on behavioral teaming and the natural computational intractability of large-scale schedule optimization suggests that robots can offer a valuable service by designing and adapting \textcolor{black}{schedules for human teammates}.

\textcolor{black}{In our system, we leverage the findings of Liu et al. to account for humans learning over time, both in problem generation as part of the \textcolor{black}{environment and} a learning curve predictor as part of the scheduling policy. The human learning curve follows an exponential function of generic form  over the course of multiple iterations as shown in Equation \ref{eq:human_learning} }\textcolor{black}{ \cite{humanrobotteam}:} \begin{equation} \label{eq:human_learning}
\textcolor{black}{y = c+ke^{-\beta i} } \end{equation} where $i$ is the number of iteration the human has previously executed a task and $c$, $k$, $\beta$  parameters. We further account for the stochastic-nature of human learning in our environment.

\subsection{Graph Neural Networks}

Graph Neural Networks (GNNs) are a class of deep neural networks that learn from unstructured data by representing objects as nodes and relations as edges and aggregating information from nearby nodes \cite{gnn_scarselli}. GNNs have been widely applied in graph-based problems such as node classification, link prediction and clustering, and \textcolor{black}{they} have shown to have an impressive performance \cite{xu2019}.
The Heterogeneous Graph Attention Network presented in Wang et al. \textcolor{black}{utilizes} Deep Learning Algorithms to address the Scheduling Problem, showing improved performance compared to non-Deep Learning Schedulers such as Earliest-Deadline First (EDF) \cite{edf_fast} and Tercio \cite{gombolay2018fast} at the cost of increased computational complexity \cite{schedulenet}.

\subsection{LSTM Based Sequence Prediction}
The impact of the LSTM network has been notable in language modeling \cite{sundermeyer2012lstm}, speech-to-text transcription\cite{graves2005bidirectional}, machine \textcolor{black}{translation \cite{zhou2021graph}}, and other applications that involve predictive modeling \cite{rnn, malhotra2016lstm, ycart2017study}. The advantage of this lengthier path generated through the recurrent nature of the neural network is that it affords an opportunity to build a certain degree of intuition that can prove beneficial during all phases of the process \textcolor{black}{ \cite{rnn, LSTMCellReview}.}

\section{HUMAN-ROBOT SCHEDULING PROBLEM}
\subsection{\textcolor{black}{Problem Overview}}


In this paper, we focus on the problem of human-robot task allocation and scheduling with temporal constraints \cite{Nunes}. We describe \textcolor{black}{the problem} components using a 4-tuple $\langle a, \tau, d, w \rangle$ \textcolor{black}{form}. $a$ represents all agents that belong to the human-robot team, $\tau$ represents all the tasks to be performed. Each task, $\tau_i$, and  agent, $a_j$, have a task completion duration $dur(\tau_i, a_j)$ and agents are capable of completing a sequence of tasks in order. $d$ contains the set of deadline constraints, where $d_i \in d$ specifies the tasks depending on $\tau_i$ \cite{schedulenet}. $w$ is the set of wait constraints where $w_{ij} \in w$ denotes the wait time between tasks $\tau_i$ and $\tau_j$. A Schedule, $S$, is a sequence of task-agent pairs $\langle \tau_i, a_j\rangle$ such that $S$ contains all tasks in $\tau$.

\subsection{Multi-Round Scheduling Environment}

The Multi-Round Scheduling Environment is developed to simulate a human-robot scheduling problem over multiple iterative rounds of execution, accounting for changes in the task performance of human workers based on previous round. Each round is a step in the OpenAI Gym-compatible environment, taking as input the complete set of task-agent pairs for the scheduling problem, simulating the sequential assignment of tasks to agents. 

Each round's execution is considered finished when all the tasks are assigned to one of the agents or if the provided schedule is determined to be infeasible under the problem constraints. The environment checks the feasibility of the provided schedule given the constraints of the problem, and computes the total duration of task completion of the schedule if the schedule is feasible. If the schedule does not satisfy the constraints, it is determined to be infeasible and the list of tasks that could not been scheduled are returned.

We formulate the Multi-Round Scheduling Environment as a \textcolor{black}{Partially Observable} Markov Decision Process (POMDP) using a six-tuple $\langle S, A, T, R, \Omega, O, \gamma\rangle$ below:
\begin{itemize}
    \item States: The problem state $S$ is a state of the Multi-Round Environment consistent of the state of the Agents.
    \item Actions: Actions at round $t$ within the Multi-Round Environment refers to a complete set of Task Allocations made up of a list of task-agent pairs, denoted as $A_t = [\langle \tau_{i_1}, a_{j_1}\rangle, \langle \tau_{i_2}, a_{j_2}\rangle, ...]$ to be executed in order. 
    \item Transitions: T corresponds to executing the action in Multi-Round Scheduling Environment and proceed to next time step.
    \item Rewards: $R_t$ is based on the scheduling objective a user wants to optimize. In Section \ref{sec:reward} we show how to compute $R_t$ when optimizing makespan.
    \textcolor{black}{
    \item Observation\textcolor{black}{s}: $\Omega$ is the estimated performance of all the task-agent pairs, plus the observable constraints.
    \item Observation Function: $O$ is handled by the \textit{Learning Curve Estimator} explained in the Section~\ref{sec:estimator}.
    }
    \item Discount factor, $\gamma$

\end{itemize}

\subsection{Agent Models}

The Multi-Round Environment stores the Agent information, allowing the environment to keep track of each agent and which tasks it has previously completed. The update of the Environment happens at the end of each round, allowing agents to modify themselves based on their internal models.  to update the model based on the selected (task-agent) pairs for each round.

\subsubsection{Determinitic Robot Model}
We generate the robot task completion times randomly through uniform distribution.
\subsubsection{Stochastic Human Model} 
We generate the human task completion times randomly based on Equation \ref{eq:human_learning}, such that the Environment can be setup to provide Deterministic and Stochastic performance for human learning. The task duration parameters of the human learning model, $c$, $k$, $\beta$, in Equation \ref{eq:human_learning} are built from the randomly selected initial task completion time for round 0. For Stochastic performance, the standard deviations are used to sample from a Normal Distribution as presented in Liu et al. \cite{humanrobotteam}.



\subsection{Learning Curve Estimator}
\label{sec:estimator}
The scheduler is given an estimate of the performance of the human agents for each task based on the information about the task duration of the previous executions of the task-agent pair through the \textcolor{black}{\textit{Learning Curve Estimator} as part of our OpenAI Gym-like Environment}
In our paper, we have implemented a black box model based on the insights presented in Liu et al\textcolor{black}{.}\cite{humanrobotteam} to simulate a Stochastic Human Learning Estimator. As an Agent completes a task in multiple rounds, the Agent Model records the task completion duration, allowing \textcolor{black}{\textit{Learning Curve Estimator}} to predict the next task-agent duration more accurately.
To represent the increase in accuracy from increase in information, we implemented a Learning Curve Estimator that generates an estimate of the human agent performance using the actual task performance as the mean of a Gaussian Distribution with noise that exponentially decreases with the number of repetitions of the same task for that agent in previous rounds.

\subsection{Reward Design}
\label{sec:reward}
The total reward, $R_t$, for the schedule generated by the multi-round scheduling environment is calculated based on feasible, $A'$, and infeasible, $\tilde A'$, subsets of task allocations, such that $A_t = A_t' \cup \tilde A_t'$. \textcolor{black}{Specifically, the reward, $R_t$, is } based on the expected reward for the feasible subset of task-agent assignments, $R_t(A_t')$,  and the reward from the assignment of the infeasible subset of task-agent assignments, $R_t{\tilde A_t'}$\textcolor{black}{...} , based on the \textcolor{black}{point estimate of the} reward from assigning the incomplete task to the agent that will complete it in the longest possible duration, multiplied by an infeasible coefficient  $C_i$ as shown in equation \ref{eq:Linearized Reward}:

\begin{equation} \label{eq:Linearized Reward}
\textcolor{black}{     R_t = \sum_{i\in A_t'}{R\left(\tau_i, a_i\right)}+ C_i \text{max}_{a_j}\left (\sum_{i \in \tilde A' }{R\left(\tau_i, a_j\right)}\right)
}
\end{equation}

\textcolor{black}{The Total Schedule Reward, $R_S$, 
favors schedules with more feasible task allocations and enables learning from infeasible explorations during training.}

\section{HYBRIDNET SCHEDULING POLICY}

As shown in \textcolor{black}{Figure} \ref{fig:pipeline}, our HybridNet framework consists of a heterogeneous graph-based encoder to learn high level embeddings of the scheduling problem, and a recurrent schedule \textcolor{black}{propagator} to generate the team schedule sequentially. This hybrid network architecture enables directly learning useful features from the problem structure, \textcolor{black}{owing} to the expressiveness of heterogeneous graph neural networks, and at the same time efficiently constructing the schedule with our LSTM-based \textcolor{black}{propagator}. As a result, HybridNet does not require interacting with the environment between every task-agent pair selection, which is necessary but computationally expensive in prior \textcolor{black}{work \cite{learning_scheduling_with_gnn, schedulenet}.}

We denote the policy learned by HybridNet as $\pi_\theta(A|S)$, with $\theta$ representing the~\textcolor{black}{parameters of the neural network}. At round $t$, an action takes the form of an ordered sequence of scheduling decisions, $A_t = \{d_1, d_2, ..., d_n\}, d_i = \langle \tau_{i}, a_{j}\rangle $, where a latter decision, $d_i$, is conditioned on its former ones, $d_{1:i-1}$. Then, the policy can be factorized as\begin{equation}
p_\theta(A_t|S_t) = \prod_{i=1}^n p_\theta(d_i|S_t, d_{1:i-1})
\label{eq:prob}
\end{equation} Using the Recurrent \textcolor{black}{Schedule Propagator}, HybridNet recursively computes the conditional probability, $p_\theta(d_i|S_t, d_{1:i-1})$, for sampling a task-agent pair. At the end, the network collects all the decisions and sends to the environment for execution.

\subsection{Heterogeneous Graph Encoder}



\begin{figure}[t]
      \centering
      \framebox{\parbox{3in}{
      \includegraphics[width=0.9\columnwidth]{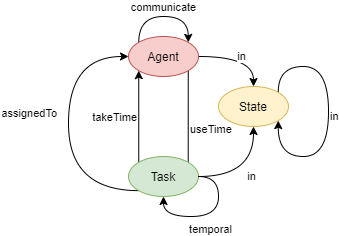}
}}
    \caption{Metagraph of the heterogeneous graph built from the STN by adding agent and state summary nodes.}
    \label{fig:gat}
\end{figure}

We build our Encoder using the heterogeneous graph attention (HetGAT) layer proposed in \cite{schedulenet} that has been shown effective in representation learning of multi-agent scheduling problems.
At the start of each round for a given human-robot scheduling problem, the heterogeneous graph representation is built by extending from the simple temporal network (STN) that encodes the temporal constraints to include agent nodes and a state summary node. The metagraph of the resulted graph is shown in Figure \ref{fig:gat}, which summarizes all the node types and edge types. Then, a HetGAT layer computes the output node features by performing per-edge-type message passing followed by per-node-type feature reduction, while utilizing a feature-dependent and structure-free attention mechanism. We refer interested readers to \cite{schedulenet} for full details of implementing a HetGAT layer.


By stacking several HetGAT layers sequentially, we construct the Encoder that utilizes multi-layer structure to extract high-level embeddings of each node that will be send to the propagator for schedule generation. \textcolor{black}{We follow the same hyper-parameters for HetGAT layers as provided in Wang et al. \cite{schedulenet}}


\subsection{Recurrent Schedule \textcolor{black}{Propagator}}

\textcolor{black}{The HetGAT layers are computationally complex and require interactive scheduling to generate the initial model. By utilizing an LSTM based Recurrent Predictor, we propagate forward consequences of each task-agent assignment, recreating the encoded information about the environment without relying on the initial HetGAT Layer, significantly \textcolor{black}{reducing} the computational complexity of our scheduler.}

The Recurrent Schedule \textcolor{black}{Propagator} takes as input the Task, State and Agent embeddings generated by the Heterogeneous Graph Encoder and sequentially generates task-agent pairs based on the encoded information. To predict the consecutive encoding of state and agents, we use an LSTM Model to recursively generate the Agent and State after each assignment of a task to an agent, without interacting with the Environment, outputting the sequential task-agent assignment for the complete set of tasks.
The pseudo-code for scheduling generation with HybridNet is presented in Algorithm~\ref{alg:schedule_production}.



\begin{algorithm}[t]
\caption{\textcolor{black}{Psuedocode for Schedule Generation}}
\textbf{Input:} graph $g$, features $f$, unscheduled-Tasks $u$\\
\textbf{Output:} \textit{schedule}
\begin{algorithmic}[1]
\label{alg:schedule_production}
\STATE $\textit{schedule} = [\ ], i = 1$
\STATE $(h_{a_1}, c_{a_1}, h_{t_1}, c_{t_1},  h_{s_1}, c_{s_1})\gets Encoder(g, f) $
\WHILE{$|\textit{u}| \neq 0$}

\STATE $p_{a_i} \gets \textit{AgentSelector}(h_{s_i}, h_{a_i})$
\STATE $a_i \gets \textit{Sampling}(p_{a_i})$
\STATE $p_{t_i} \gets \textit{TaskSelector}(h_{t_i}, h_{s_i}, a_i)$
\STATE $t_i \gets \textit{Sampling}(p_{t_{i-1}})$
\STATE $\textit{schedule.append}(\langle t_i, a_i \rangle)$
\STATE $\textit{unscheduledTasks.remove}(t_i)$
\IF{$|\textit{unscheduledTasks}| == 0$}
\STATE \textbf{return} \textit{schedule}
\ENDIF
\STATE $i \gets i + 1$ \\
\STATE $h_{s_i}, c_{s_i} \gets LSTM_s((h_{t_{i-1}}[t_i],h_{a_{i-1}}[a_i]),$ \\
\hspace*{30mm} $h_{a_{i-1}}, c_{a_{i-1}})$ \\
 \STATE $h_{a_i}, c_{a_i} \gets LSTM_a((h_{t_{i-1}}[t_i],h_{a_{i-1}}[a_i]), $\\
 \hspace*{30mm} $h_{a_{i-1}}, c_{a_{i-1}})$ \\
\ENDWHILE
\end{algorithmic}
\vspace{-6mm}
\end{algorithm}

As $d_i = \langle \tau_{i}, a_{j}\rangle $, we further factor $p_\theta(d_i|S_t, d_{1:i-1})$ into an agent selector and a task selector. That is, $\pi_{factor}(d|\cdot) = \pi_{agent}(a_j|\cdot) \cdot \pi_{task}(\tau_i|a_j, \cdot)$. This factorization allows the policy to capture the underlying composite and conditional nature of the scheduling decisions, where the task to schedule is strongly dependent on the picked agent.


The Agent Selector selects the new agent for the next decision $d$ based on the state and agent information. Specifically, the concatenated state-agent embeddings are processed by a feed-forward neural network, $f_a$, to compute the likelihood of selecting each agent for the next task-agent pair, using Equation \ref{eq:softmax_action}. A softmax operation is performed to convert the raw predictions into a probability distribution. After the selection of the agent, \textcolor{black}{the agent} embedding of the chosen agent is updated based on the selected task and state embeddings, as state change only happens for the assigned agent. This approach allows the agent selector to consider how busy each agent is, based on the inherent information presented in the embeddings.
\vspace{-1mm}
\begin{equation}
\vspace{-1mm}
    \label{eq:softmax_action}
    \pi_{agent}(a_j|s) = \textit{softmax}_i(f_a( [h_{a_j} || h_s]))
\end{equation}

Next, the \textcolor{black}{Schedule Propagator} uses the Task Selector to assign the task for the selected agent based on the state, agent and unscheduled task embeddings. As shown in Equation \ref{eq:softmax_task}, the Task Selector concatenates the state, selected agent and the unscheduled task embeddings and passes the combined information to a feedforward neural network, $f_\tau$, to calculate the likelihood of the task being assigned to the selected agent. After assigning to an agent for execution, the tasks are removed from the list of unscheduled tasks. Since the calculation of likelihood of each task is independent of each other up to the last softmax operation, the model is scalable and can be used for different\textcolor{black}{ problem sizes.}
\vspace{-1mm}
\begin{equation}
\vspace{-1mm}
    \label{eq:softmax_task}
    \pi_{task}(\tau_i|a_j, s) = \textit{softmax}_i(f_\tau( [h_{\tau_i} || h_{a_j} || h_s ]))
\end{equation}


The key component of the \textcolor{black}{Schedule Propagator} is the use of LSTM. As shown in line 12 of Algorithm 1, after each task-agent pair selection, the state and agent embeddings are updated using the state LSTM and agent LSTM, respectively. The LSTM Cell stores the hidden and cell data from the previous step of the task allocation and \textcolor{black}{predicts} the next step based on the input using the 
Equation \ref{eq:LSTM_Update} \cite{LSTMCellReview}.
\vspace{-2mm}
\begin{equation}
    \label{eq:LSTM_Update}
    \begin{split}
    &f_t = \sigma(W_{f}[h_{t-1}, x_t] + b_{f})\\
    &i_t = \sigma(W_{i}[h_t, x_t] + b_i) \\
    &\tilde c_t = \textit{tanh}(W_c[h_{t-1}, x_t] + b_c) \\
    &c_t = f_t c_{t-1} + i_t \tilde c_t \\
    &o_t = \sigma(W_{o} [h_{t-1}, x_t]+ b_{o}\\
    &h_t = o_t \textit{tanh}(c_t)
    \end{split}
\end{equation} Where the Encoder produces initial hidden state, $h_1$ and initial cell state $c_1$ as an output in the form of $[h_1, c_1]$. 

\textcolor{black}{During testing, we utilize a batched sampling strategy for further performance gains. Specifically, we generate multiple schedules for the same task allocation problem every round. We select the best performing schedule by computing the estimated makespan utilizing the Learning Curve Estimator and provide it to the Multi-Round Environment. More sampling improves solution quality at increased computation.
}

\subsection{Stochastic Policy Learning}

We train HybridNet in multi-round scheduling environments using Policy Gradient methods that seek to directly optimize the model parameters based on rewards received from the environment \textcolor{black}{\cite{sutton2000comparing}}. Specifically, we compute the gradient of the model using the sum of the log likelihood of Agent and Task Selectors, as shown in Equation
\vspace{-2mm}
\ref{eq:Monte Carlo}:
\begin{equation}
    \label{eq:Monte Carlo}
    \begin{split}
    \nabla_\theta J(\theta) = \mathbb{E}_\pi (\sum_t^T&A_t^{\pi_\theta}(s_t, \langle \tau_i, a_i\rangle ) \\
    &\nabla_\theta (log\pi_\theta(\tau_i | a_i, s_t) + log\pi_\theta(a_i | s_t))
    \end{split}
\end{equation}

In Equation \ref{eq:Monte Carlo}, the advantage term, $A_t$ is estimated by subtracting a \textcolor{black}{“baseline"} from the total future reward calculated in Equation \ref{eq:Linearized Reward}. We calculate the \textcolor{black}{“baseline"} using the reward generated for the same task-allocation problem from multiple batches executed in multiple sequential rounds in the Multi-Round Environment. Each element of the batch solves the same scheduling problem and the environment is updated to account for the task-allocation of the previous round, updating the agent models. 
\textcolor{black}{The gradients were calculated from Equation~\ref{eq:Monte Carlo} to updated the model weights}.

Due to the combinatorial nature of the task scheduling problem, plus the stochasticity in human proficiency, learning a helpful value function as a baseline for computing the advantage term is non-trivial. Instead, we investigate two more accessible and efficient alternatives:
\begin{itemize}
    \item \emph{Step-based Baseline:}
    \textcolor{black}{During gradient estimation, the baseline value subtracted is set as the average return value across training episodes in the current batch.}

    \item \emph{Greedy Rollout Baseline:}
    Greedy Rollout Baseline uses, $\pi_{greedy}(A|S)$, a deterministic greedy version of the HybridNet scheduler, to collect rewards in the environment. Its weights, $\theta_{greedy}$, are updated periodically by copying the weights from the current learner, $\pi_\theta(A|S)$.
    


\end{itemize}



\section{EXPERIMENTAL RESULTS}

\subsection{Data Generation}

\textcolor{black}{We generate scheduling problems with deadline and wait constraints under different scales. For all scales, the deadline constraints are randomly generated for approximately 25\% of the tasks from a range of [1, $5N$]} where $N$ is the number of tasks. Approximately 25\% of the tasks have wait constraints, and the duration of non-zero wait constraints is sampled from \textcolor{black}{$U([1, 10])$. Task durations are clamped to 10 to 100.}


\subsubsection{Small Scale}
The small data set \textcolor{black}{has 9 to 11 tasks with 2 robots and 2 humans in a team. We generated 2000 Training Problems and 200 Test Problems.}

\subsubsection{Medium Scale}
The medium data set \textcolor{black}{has 18 to 22 tasks with 2 robots and 2 humans in a team. We generated 2000 Training Problems and 200 Test Problems}
\textcolor{black}{to inspect the scalability of our trained model.}

\subsubsection{Large Scale}
The large data set is defined as problems with 36 to 44 tasks chosen at random with 2 robots and 2 humans in a team. We have generated 200 Test Problems to evaluated the \textcolor{black}{HybridNet performance with zero training problems (i.e., zero-shot transfer to from the smaller scale datasets to the Large Scale dataset).} 

\textcolor{black}{To simulate the stochastic learning of human agents, for each Data Set noise is introduced to the Human Agent models by simulating the natural distribution of the $c$, $k$, $\beta$  parameters of Equation \ref{eq:human_learning}. This allows for each Data Set to simulate Deterministic and Stochastic Human Performance. The stochastic model is clipped to fall within the specified range of task durations.}
\subsection{Benchmarking}
\label{section:Benchmarking}
We benchmark HybridNet against the following methods:
\begin{itemize}
    \item \emph{EDF}: A ubiquitous heuristic algorithm, earliest deadline first (EDF), that selects from a list of available tasks
the one with the earliest deadline, assigning it to the first
available agent.
    \item \emph{Genetic Algorithm}: An Evolutionary Optimization Algorithm that uses Post-Processing on the Schedule Generated by EDF \cite{Zhang2020RealTimeAA}. Genetic algorithm creates new schedules based on the initial schedule through iterative randomized mutations by swapping task allocations and task orders \cite{humanrobotteam}. Each generation selects the top performing schedules, sorted on feasibility and total schedule completion time, and used as the baseline for creating new mutations. The Genetic Algorithm was run for 10 generation with 90 baseline schedules, 10 task allocation and 10 task order swapping mutations.
\end{itemize}

\textcolor{black}{Furthermore, we evaluate the functionality of the Recurrent Schedule Propagator by comparing it against the following HybridNet variant:}
\begin{itemize}
    \item \textcolor{black}{\emph{HetGAT}: We implement a HetGAT Scheduler based on the Encoder of HybridNet. After each task-agent pair assignment, instead of using the LSTM Cells to update the task, agent and state embeddings, it directly interacts with the environment to model the consequences of action with a new heterogeneous graph and re-computes those information from it.
    }
\end{itemize}
\begin{table*}[h]
\caption{Evaluation Results: Adjusted Makespan and Feasibility with Deterministic Human Task Proficiency \textcolor{black}{comparing Benchmarks with HybridNet trained on Small and Medium scales, with schedules sampled from sizes 8 and 16}}
\vspace{-5mm}
\begin{center}
\resizebox{.95\textwidth}{!}{
\begin{tabular}{llllllll}
\hline
\multirow{2}{*}{\textbf{Training}}   & \multirow{2}{*}{\textbf{Methods}} & \multicolumn{2}{c}{\textbf{Small}}                       & \multicolumn{2}{c}{\textbf{Medium}}                    & \multicolumn{2}{c}{\textbf{Large}}                      \\
                                     &                                   & \textbf{Makespan}           & \textbf{Feasibility (\%)}  & \textbf{Makespan}          & \textbf{Feasibility (\%)} & \textbf{Makespan}           & \textbf{Feasibility (\%)} \\ \hline
\textbf{-}                           & \textbf{EDF}                      & 239.31                      & 73.00                      & 1109.85                    & 15.00                     & 2535.89                     & 1.00                      \\
\textbf{-}                           & \textbf{Genetic Algorithm}        & 302.42  $\pm$ 0.77          & 74.10  $\pm$ 0.30          & 1180.07 $\pm$2.54          & 16.60  $\pm$ 0.70         & 2542.79 $\pm$ 0.06          & 1.00  $\pm$ 0.00          \\
\multirow{2}{*}{\textbf{Step-based}} & \textbf{HetGAT 8}                 & 257.20  $\pm$ 0.18          & 86.29 $\pm$ 0.08           & 751.27  $\pm$ 1.29         & 50.17  $\pm$ 0.14         & 2123.96  $\pm$ 5.66         & 17.12  $\pm$ 0.27         \\
                                     & \textbf{HetGAT 16}                & 249.69 $\pm$ 0.30           & 86.51  $\pm$ 0.09          & 723.57  $\pm$ 0.94         & 50.29  $\pm$ 0.11         & 2081.65  $\pm$ 5.45         & 17.15  $\pm$ 0.16         \\
\multirow{2}{*}{\textbf{Greedy}}     & \textbf{HetGAT 8}                 & 261.15 $\pm$ 0.09           & 85.59 $\pm$ 0.10           & 784.32  $\pm$ 0.52         & 53.28  $\pm$ 0.17         & 2017.25  $\pm$ 2.16         & 23.98  $\pm$ 0.14         \\
                                     & \textbf{HetGAT 16}                & 255.70 $\pm$ 0.23           & 86.05  $\pm$ 0.15          & 765.79  $\pm$ 0.96         & 53.41  $\pm$ 0.08         & 1983.73  $\pm$ 1.59         & 23.84  $\pm$ 0.01         \\ \hline
\multirow{2}{*}{\textbf{Step-based}} & \textbf{HybridNet Small 8}        & 260.22  $\pm$ 0.15          & 86.93  $\pm$ 0.10          & 770.48  $\pm$ 1.07         & 59.11  $\pm$ 0.35         & 2005.80  $\pm$ 2.33         & 30.65  $\pm$ 0.39         \\
                                     & \textbf{HybridNet Small 16}       & \textbf{252.57  $\pm$ 0.49} & \textbf{87.08  $\pm$ 0.10} & 746.35  $\pm$ 0.52         & 60.89  $\pm$ 0.36         & 1953.65  $\pm$ 3.76         & 33.24  $\pm$ 0.61         \\
\multirow{2}{*}{\textbf{Greedy}}     & \textbf{HybridNet Small 8}        & 266.74  $\pm$ 0.31          & 84.65  $\pm$ 0.32          & 758.96 $\pm$ 2.27          & 61.09  $\pm$ 0.43         & 2049.32  $\pm$ 3.73         & 28.74  $\pm$ 0.45         \\
                                     & \textbf{HybridNet Small 16}       & 258.17  $\pm$ 0.45          & 85.13  $\pm$ 0.20          & 723.35 $\pm$ 1.70          & 63.68 $\pm$ 0.49          & 1973.15 $\pm$ 2.91          & 32.46 $\pm$ 0.40          \\
\multirow{2}{*}{\textbf{Step-based}} & \textbf{HybridNet Medium 8}       & -                           & -                          & 722.85 $\pm$ 0.61          & 64.69 $\pm$ 0.29          & 2010.86 $\pm$ 1.97          & 30.86 $\pm$ 0.45          \\
                                     & \textbf{HybridNet Medium 16}      & -                           & -                          & 697.40 $\pm$ 2.04          & 66.25 $\pm$ 0.51          & 1944.72 $\pm$ 4.10          & 33.88 $\pm$ 0.49          \\
\multirow{2}{*}{\textbf{Greedy}}     & \textbf{HybridNet Medium 8}       & -                           & -                          & 692.01 $\pm$ 3.69          & 68.33 $\pm$ 0.66          & 2011.78 $\pm$ 5.08          & 30.58 $\pm$ 0.87          \\
                                     & \textbf{HybridNet Medium 16}      & -                           & -                          & \textbf{659.01 $\pm$ 0.89} & \textbf{71.00 $\pm$ 0.45} & \textbf{1936.97 $\pm$ 4.68} & \textbf{34.66 $\pm$ 0.74} \\ \cline{2-8} 
\end{tabular}
}
\end{center}
\label{table:performance}
\vspace{-2mm}
\end{table*}

\begin{table*}[h]
\vspace{-2mm}
\caption{Evaluation Results: Adjusted Makespan and Feasibility with Stochastic Human Task Proficiency}
\vspace{-5mm}
\begin{center}
\resizebox{.95\textwidth}{!}{
\begin{tabular}{lllllll}
\hline
\multirow{2}{*}{\textbf{Methods}} & \multicolumn{2}{c}{\textbf{Small}}                     & \multicolumn{2}{c}{\textbf{Medium}}                    & \multicolumn{2}{c}{\textbf{Large}}                      \\
                                  & \textbf{Makespan}          & \textbf{Feasibility (\%)} & \textbf{Makespan}          & \textbf{Feasibility (\%)} & \textbf{Makespan}           & \textbf{Feasibility (\%)} \\ \hline
\textbf{EDF}                      & 227.81$\pm$ 6.17           & 75.65 $\pm$ 1.21          & 1071.02$\pm$ 20.65         & 17.30 $\pm$ 1.12          & 2524.92$\pm$ 8.95           & 1.15 $\pm$ 0.23           \\
\textbf{Genetic Algorithm}        & 283.79 $\pm$ 10.39         & 77.45 $\pm$ 2.05          & 1149.42 $\pm$ 12.14        & 19.55 $\pm$ 1.31          & 2541.20 $\pm$ 3.54          & 1.05 $\pm$ 0.15           \\
\textbf{HybridNet Small}          & \textbf{298.81 $\pm$ 0.96} & \textbf{79.54 $\pm$ 0.52} & 881.16 $\pm$ 2.89          & 48.89 $\pm$ 1.09          & \textbf{2141.80 $\pm$ 5.12} & \textbf{23.51 $\pm$ 0.96} \\
\textbf{HybridNet Medium}         & -                          & -                         & \textbf{859.99 $\pm$ 4.82} & \textbf{51.94 $\pm$ 1.32} & 2174.57 $\pm$ 8.53          & 22.31 $\pm$ 0.94          \\ \hline
\end{tabular}
}
\end{center}
\label{table:noisy_performance}
\vspace{-7mm}
\end{table*}

\begin{table}[h]
\caption{Evaluation Results: \textcolor{black}{Runtime Performance on Single Problem}}
\vspace{-5mm}
\begin{center}
\resizebox{.95\columnwidth}{!}{
\begin{tabular}{lllll}
                                              & \textbf{Methods} & \textbf{HetGAT8}   & \textbf{HybridNet8}       & \textbf{HybridNet16} \\ \hline
\multirow{2}{*}{\textbf{Training Time (s)}}   & \textbf{Small}   & 184.52 $\pm$ 18.00 & \textbf{19.97 $\pm$ 0.91} & -                    \\
                                              & \textbf{Medium}  & 354.77 $\pm$ 38.31 & \textbf{22.40 $\pm$ 6.52} & -                    \\ \hline
\multirow{3}{*}{\textbf{Evaluation Time (s)}} & \textbf{Small}   & 22.91   $\pm$ 5.85 & \textbf{10.94 $\pm$ 0.99} & 18.95 $\pm$ 3.53     \\
                                              & \textbf{Medium}  & 70.12 $\pm$ 8.67   & \textbf{14.77 $\pm$ 1.42} & 22.30 $\pm$ 7.55     \\
                                              & \textbf{Large}   & 123.76 $\pm$ 32.32 & \textbf{18.84 $\pm$ 7.38} & 27.78 $\pm$ 16.52    \\ \cline{2-5} 
\end{tabular}
}
\end{center}
\label{table:performance_time}
\vspace{-7mm}
\end{table}

We evaluate HybridNet on three metrics: 1) Proportion of
problems solved; 2) Adjusted makespan: determined by the average of the makespan of feasible schedules and the maximum possible makespan of the infeasible schedules; and 3) Runtime statistics.
Runtime statistics for training and execution is compared for HybridNet and \textcolor{black}{HetGAT Scheduler} to model their computational complexity.
\textcolor{black}{Because HetGAT Scheduler relies on interactive scheduling through the environment after every task-agent pair allocation, we only train and evaluate it for Deterministic Human Performance.}

\subsection{Model Details}
We implement HybridNet and \textcolor{black}{HetGAT} using PyTorch \cite{paszke2019} and Deep Graph Library \cite{wang2020deep}. The HybridNet \textcolor{black}{Encoder} used in training/testing is constructed by stacking three multi-head HetGAT layers (the first two use concatenation, and the last one uses averaging). The feature dimension of hidden layers = 64, and the number of heads = 8. The Recurrent Propagator utilizes a LSTMCell of size 32 followed by a fully-connected layer and a softmax layer. We set $\gamma$ = 0.99, batch size = 8 and used Adam optimizer \cite{kingma2017adam} with a learning rate of $2\times10^{-3}$, and a weight decay of $5\times10^{-4}$. We employed a learning rate decay of $0.5$ every 4000 epochs. \textcolor{black}{We evaluate the models using a batch size of 8 and 16.} For the Multi-Round Environment, the infeasible reward coefficient $C_i=2.0$ and total round number = 4. Both training and evaluation were conducted on a Quadro RTX 8000 GPU.

\subsection{Evaluation Results}
Table~\ref{table:performance} shows the evaluation performance with Deterministic Human Proficiency in different scales. The Deterministic Human Proficiency means that during training and evaluation, human learning curve is known and execution is deterministic for every agent.
In Table~\ref{table:performance}, ``Small'' and ``Medium'' term after model name denotes the data scale the model was trained on and the number following it denotes the batch size for schedule sampling.
The results show that HybridNet outperforms both EDF and Genetic Algorithm in adjusted makespan and percentage of feasibility.
HybridNet trained on Small scale problems generalizes for both Medium and Large scale problems with similar or slightly worse performance than HybridNet trained on Medium.
HybridNet and HetGAT performs similarly on all scales. This shows that HybridNet is capable of learning high performance policies by leveraging the Recurrent Schedule Propagator and without requiring interaction with the Environment.

We provide the runtimes of training and evaluation for HetGAT and HybridNET in Table~\ref{table:performance_time}. 
HybridNet is approximately 10 times faster in training compared to HetGAT Model and \textcolor{black}{at least} 2 times faster during evaluation for same batch size. EDF and Genetic Algorithm were evaluated through the CPU without GPU acceleration, making it infeasible to accurately compare the performance of the Deep Learning Models to the Traditional Models. 

We show that for HybridNet, step-based training has better performance over the greedy baseline, while for HetGAT model, greedy baseline training is better. We also observed that greedy baseline training reached convergence faster than step-based training (4500 epochs vs. 19000 epochs).
\textcolor{black}{Further investigation is worthwhile.} 

Table \ref{table:noisy_performance} shows the evaluation performance with Stochastic Human Proficiency in different scales. The Stochastic Human Proficiency is presented as randomness in both the actual human execution within Multi-Round Environment and uncertainty within the Learning Curve Estimator used for schedule generation. The results show that HybridNet outperforms the EDF and Genetic Algorithm across different data scales. The largest performance gap was observed on large dataset \textcolor{black}{(23.51\% vs. 1.15\%)}.
Here, HetGAT model is not included as it requires interaction with the environment after every task-agent assignment to observe the outcome, which is not available until the whole schedule is generated and sent to the Stochastic Environment for execution to emulate \textcolor{black}{real-world} scenarios.





\section{CONCLUSIONS}

We present a deep learning-based framework, called HybridNet, combining a heterogeneous graph-based encoder with a recurrent schedule propagator, for scheduling stochastic human-robot teams under temporal constraints.
The \textcolor{black}{resulting} policy network provides a computationally lightweight yet highly expressive model that is end-to-end trainable via reinforcement learning algorithms.
We developed a multi-round task scheduling environment for stochastic human-robot teams and conducted extensive experiments, showing that HybridNet outperforms other human-robot scheduling solutions across problem sizes. 
Future research includes integrating the learning-based human estimator into HybridNet, transfer learning across optimizing different objective functions, and deploying the trained network in a real-world scenario.


\vspace{-2mm}

\bibliographystyle{IEEEtran}
\bibliography{IEEEabrv, IEEE_IROS_HybridNET}

\end{document}